\documentclass[conference]{IEEEtran}
\IEEEoverridecommandlockouts

\usepackage{cite}
\usepackage{amsmath,amssymb,amsfonts}
\usepackage{algorithmic}
\usepackage{graphicx}
\usepackage{textcomp}
\usepackage{xcolor}
\usepackage{booktabs}
\usepackage{multirow}
\usepackage{diagbox}
\usepackage[backref]{hyperref}
\hypersetup{
hidelinks}
\ifCLASSOPTIONcompsoc
\usepackage[caption=false,font=normalsize,labelfon
t=sf,textfont=sf]{subfig}
\else
\usepackage[caption=false,font=footnotesize]{subfi
g}
\fi
\def\BibTeX{{\rm B\kern-.05em{\sc i\kern-.025em b}\kern-.08em
    T\kern-.1667em\lower.7ex\hbox{E}\kern-.125emX}}
\begin{document}

\title{Understanding Robustness of Parameter-Efficient Tuning for Image Classification}

\author{
\IEEEauthorblockN{Jiacheng Ruan, Xian Gao, Suncheng Xiang, Mingye Xie, Ting Liu and Yuzhuo Fu}
\IEEEauthorblockA{Shanghai Jiao Tong University \\
Email: \{jackchenruan, gaoxian, xiangsuncheng17, xiemingye, louisa\_liu, yzfu\}@sjtu.edu.cn}
}

\maketitle

\begin{abstract}
Parameter-efficient tuning (PET) techniques calibrate the model's predictions on downstream tasks by freezing the pre-trained models and introducing a small number of learnable parameters. However, despite the numerous PET methods proposed, their robustness has not been thoroughly investigated. In this paper, we systematically explore the robustness of four classical PET techniques (e.g., VPT, Adapter, AdaptFormer, and LoRA) under both white-box attacks and information perturbations. For white-box attack scenarios, we first analyze the performance of PET techniques using FGSM and PGD attacks. Subsequently, we further explore the transferability of adversarial samples and the impact of learnable parameter quantities on the robustness of PET methods. Under information perturbation attacks, we introduce four distinct perturbation strategies, including Patch-wise Drop, Pixel-wise Drop, Patch Shuffle, and Gaussian Noise, to comprehensively assess the robustness of these PET techniques in the presence of information loss. Via these extensive studies, we enhance the understanding of the robustness of PET methods, providing valuable insights for improving their performance in computer vision applications. The code is available at \href{https://github.com/JCruan519/PETRobustness}{https://github.com/JCruan519/PETRobustness}.

\end{abstract}


\section{Introduction}
\label{sec.intro}

With the continuous increase in the number of model parameters \cite{vit22b,vitsurvey,llama2,llamamoe,llama3}, the cost of fine-tuning pretrained models on downstream tasks using full parameters is gradually rising. Recently, a technique known as Parameter-Efficient Tuning (PET) \cite{peftreview,bitfit,compacter,maple,nlpprefix,nlpprompt,Neuralpromptsearch,ssf} has emerged. This approach involves freezing the backbone of the model and using a small number of learnable parameters to adjust the model's predictions on downstream tasks, thereby achieving performance comparable to those of full parameter fine-tuning.

The existing PET methods can be categorized into Prompt Tuning, Adapter Tuning, and Parameter Tuning \cite{VisualTuning}. As a representative of Prompt Tuning, Visual Prompt Tuning (VPT) \cite{VPT} introduces learnable tokens to acquire downstream knowledge. Adapter \cite{nlpadapter,adaptformer}, representing Adapter Tuning, adjust the model's output distribution by adding new bottleneck layer modules. LoRA \cite{nlplora}, representing Parameter Tuning, calibrates the model's predictions through low-rank approximation of the update matrices. Subsequent studies have largely focused on improving these classic PET techniques and have seen extensive development in the fields of vision \cite{convpass,readapter,gist,idat}, language \cite{attempt,compacter}, and multimodal \cite{clipadapter,maple,lamm}. Despite notable successes, the robustness of PET techniques has not been thoroughly explored.

In this paper, we systematically analyze the robustness of PET techniques on the VTAB-1K benchmark \cite{vtab1k} for the first time. VTAB-1K is a challenging image classification dataset that comprises 19 distinct subtasks. The PET techniques examined in our study include four classic methods\footnote{In this paper, Adapter \cite{nlpadapter} is referred to as S-Ada., and AdaptFormer \cite{adaptformer} is referred to as P-Ada.}: VPT \cite{VPT}, S-Ada. \cite{nlpadapter}, P-Ada. \cite{adaptformer} and LoRA \cite{nlplora}. The attack scenarios include two types of white-box attacks and four types of information perturbations.

Under white-box attacks, we employ the Fast Gradient Sign Method (FGSM) \cite{FGSM} and Projected Gradient Descent (PGD) \cite{PGD}. Initially, we compare the robustness differences between traditional fine-tuning methods, such as Linear Probing (LP), and the PET techniques. Subsequently, we investigate the transferability of adversarial samples. Finally, we analyze the impact of varying learnable parameter quantities on the robustness of PET techniques. The findings reveal that: \textbf{1)} LP demonstrates superior adversarial robustness compared to PET techniques, with VPT and LoRA performing relatively well within PET methods; \textbf{2)} LoRA exhibits the best robustness accuracy when facing adversarial samples from different sources, and the transferability of adversarial samples between similar PET techniques has a significant impact on robustness; \textbf{3)} PET methods vary in their sensitivity to changes in the number of learnable parameters, with VPT being the least sensitive, LoRA being more sensitive, and S-Ada. and P-Ada. showing moderate sensitivity.

In information perturbation attacks, we adopt four different perturbation methods and conduct experiments under varying levels of perturbation intensity. The results indicate that Patch-wise Drop and Patch Shuffle primarily disrupt the ViT architecture with minimal association to specific fine-tuning methods. In contrast, Pixel-wise Drop and Gaussian Noise, introducing sparse information loss and noise to each image token, have a greater impact on VPT, while their effects on the other three PET techniques are relatively minor.

\section{Preliminaries}
\label{sec.rela}

\textbf{VPT} draws inspiration from the Prompting in natural language processing \cite{nlpprompt}. It involves concatenating a learnable sequence \(P \in \mathbb{R}^{p \times d}\) with the image sequence \(I \in \mathbb{R}^{n \times d}\) to form a new input sequence \(X \in \mathbb{R}^{(p+n) \times d}\) for processing by ViT \cite{vit} blocks. VPT has two variants: VPT-Shallow and VPT-Deep, where the former adds \(P\) only before the first ViT block, while the latter introduces different \(P\) at each block.

In a Transformer block \cite{TRM,vit}, \textbf{S-Ada.} employs a bottleneck structure, sequentially inserted after the multi-head self-attention module (MHSA) and the feed-forward neural network module (FFN), expressed as equation \ref{eq:adapter}. This structure includes a down projection layer \(W_d \in \mathbb{R}^{d \times d'}\), an activation function \(\sigma\), and an up projection layer \(W_u \in \mathbb{R}^{d' \times d}\). Furthermore, \textbf{P-Ada.} modifies the sequential insertion into a parallel configuration, attaching it alongside the bypass of the FFN, and introduces a scaling factor, expressed as equation \ref{eq:adaptformer}.

\begin{equation}
\begin{aligned}
    X^{\prime} &= \sigma\left(\operatorname{MHSA}\left(\operatorname{LN}(X)\right) W_d\right) W_u + X, \\
    Y &= \sigma\left(\operatorname{FFN}\left(\operatorname{LN}(X^{\prime})\right) W_d\right) W_u + X^{\prime}
\end{aligned}
\label{eq:adapter}
\end{equation}

\begin{equation}
    Y = \operatorname{FFN}\left(\operatorname{LN}(X^{\prime})\right) + s \cdot \sigma\left(X^{\prime} W_d\right) W_u + X^{\prime}
    \label{eq:adaptformer}
\end{equation}
where \( s \) is the scaling factor, and $\operatorname{LN}$ represents Layer Normalization \cite{layernorm}. It is worth noting that when \( d' \ll d \), the model can be fine-tuned using a very small number of parameters.

The key to \textbf{LoRA} lies in the low-rank approximation of the updated weight matrices. Specifically, LoRA introduces low-rank matrices \( \Delta W_{q/v} \) on \( W_{q/v} \) in the MHSA. Thus, the computation for query and value can be expressed as follows.

\begin{equation}
    \Delta W_{q/v} = s \cdot A_{q/v} B_{q/v}; \quad Q/V = X \left(W_{q/v} + \Delta W_{q/v}\right)
\end{equation}
where \( A_{q/v} \in \mathbb{R}^{d \times r} \) and \( B_{q/v} \in \mathbb{R}^{r \times d} \) represent the low-rank matrices, and \( s \) denotes the scaling factor. Similarly, when \( r \ll d \), the amount of learnable parameters introduced by LoRA is also very small.

\begin{table*}[!t]
\caption{Accuracy (\%) on the three groups of VTAB-1K. $\Delta$ stands for the gap between White-box attacks and Clean results.}
\label{tab:white_box_att}
\centering
\setlength{\tabcolsep}{4pt}
\begin{tabular}{lc|ccc|cccccc|cccccc}
\toprule
\multirow{2}{*}{Method} & \multirow{2}{*}{Params. (M)} & \multicolumn{3}{c|}{Clean} & \multicolumn{6}{c|}{FGSM} & \multicolumn{6}{c}{PGD}  \\ 
 & &Nat.&Spe.&Str. &Nat. &$\Delta$ & Spe. &$\Delta$ &Str. &$\Delta$ &Nat. &$\Delta$ &Spe. &$\Delta$ &Str. &$\Delta$ \\
\midrule
LP &0.04 
&70.43&72.25&35.90 
&41.11&29.31&35.00&37.25&5.03&30.88
&33.83&36.60&39.55&32.70&5.13&30.78 \\
VPT &0.06 
&70.83&72.60&43.08 
&39.91&30.91&34.70&37.90&11.10&31.98  
&34.74&36.09&40.90&31.70&9.25&33.83 \\
S-Ada. &0.17 
&78.54&73.85&54.50 
&47.26&31.29&36.05&37.80&19.53&34.98
&41.34&37.20&40.45&33.40&17.98&36.53 \\
P-Ada. &0.17 
&79.00&73.65&57.25
&51.14&27.86&34.45&39.20&17.60&39.63
&43.17&35.83&37.35&36.30&17.18&40.08 \\
LoRA &0.15 
&79.26&72.75&55.10 
&49.66&29.60&36.10&36.65&17.10&38.00
&42.29&36.97&40.30&32.45&18.93&36.18 \\
\bottomrule
\end{tabular}
\end{table*}

\begin{table}[!t]
\caption{Transferability of White-box Attacks on VTAB-1K benchmark. Mean Accuracy (\%) across 19 datasets are reported. PGD is utilized as the attack method.}
\label{tab:tras_white_box_attack}
\centering
\begin{tabular}{c|c|c|c|c|c}
\toprule
\diagbox{Source}{Target} &LP  &VPT &S-Ada.  &P-Ada.  &LoRA \\
\hline
LP &22.95  &37.06  & 52.84 & 49.81 & 53.07 \\
\hline
VPT & 29.87 & 25.31 & 52.44 & 49.85 & 53.04 \\
\hline
S-Ada. & 36.66 & 43.25 & 31.32 & 40.71 & 49.14 \\
\hline
P-Ada. & 32.29 & 39.85 & 40.05 & 31.00 & 47.75 \\
\hline
LoRA & 37.27 & 44.27 & 49.98 & 49.41 &32.03  \\
\bottomrule
\end{tabular}
\end{table}

\section{Robustness to White-box Attacks}

This section aims to evaluate the robustness of different PET techniques under white-box attacks. We select two classical white-box attack methods: FGSM \cite{FGSM} and PGD \cite{PGD}, and conduct attack experiments on four PET techniques (e.g., VPT \cite{VPT}, S-Ada. \cite{nlpadapter}, P-Ada. \cite{adaptformer}, and LoRA \cite{nlplora})\footnote{Unless otherwise specified, we default to using VPT-Shallow with 20 learnable tokens, $d^{\prime}=8$ for Adapter (S-Ada.), $s=0.1$ and $d^{\prime}=8$ for AdaptFormer (P-Ada.), and $s=1.0$ and $r=8$ for LoRA. The ViT-B/16 \cite{vit}, pre-trained on the ImageNet-21K dataset \cite{imagenet}, is employed as the backbone.}. Additionally, we investigate the transferability of adversarial samples across different PET techniques. Finally, we explore how the robustness of PET techniques is affected by varying the amount of learnable parameters. 

\subsection{Implementation Details}

We validate PET techniques on the VTAB-1K benchmark \cite{vtab1k}. Specifically, VTAB-1K is a benchmark for image classification tasks that includes 19 different datasets, categorized into three groups: Natural (Nat.), Specialized (Spe.), and Structured (Str.). For each dataset, we train using 1,000 samples and evaluate on 500 samples. For both FGSM and PGD, we apply an $l_{\infty}$-norm with a perturbation magnitude of $\epsilon=0.01$. FGSM operates in a single step, while PGD, as a multi-step extension of FGSM, iterates for 5 steps with a step size of 5e-3. In addition to PET techniques, we introduce Linear Probing (LP) as a baseline comparison, which only updates the linear classifier during fine-tuning. All experiments are conducted on NVIDIA A100 GPUs.

\subsection{White-box Attacks}
\label{sec:wba}

The results of the white-box attacks are shown in Table \ref{tab:white_box_att}. Based on these results, we can draw the following conclusions: \textbf{1)} The adversarial robustness of LP is superior to that of PET techniques; \textbf{2)} Among PET techniques, VPT and LoRA exhibit relatively better adversarial robustness, whereas S-Ada. and P-Ada. perform less effectively.

VPT introduces 20 learnable tokens, and the learnable parameters are of the same order as LP. However, under FGSM attacks, VPT shows a more significant drop in accuracy compared to LP, with an average decrease of 1.12\% across three image classification task groups. Under PGD attacks, VPT experiences a more pronounced accuracy drop on Str. group, decreasing by 3.05\% compared to LP. These differences suggest that the additional learnable tokens may be more vulnerable to adversarial attacks.

S-Ada. and P-Ada. exhibit different levels of adversarial robustness under white-box attacks. S-Ada. introduces the adapter module into the pre-trained model in a serial manner, while P-Ada. adopts a parallel insertion approach. The results indicate that the parallel insertion method shows significantly weaker adversarial robustness. Under FGSM and PGD attacks, S-Ada. outperforms P-Ada. by 0.87\% and 1.69\%, respectively, across the three task groups.

\subsection{Transferability of White-box Attacks}
\label{sec:twba}

In this section, we comprehensively evaluate the transferability of adversarial samples between different fine-tuning techniques, using PGD as the white-box attack method. In Table \ref{tab:tras_white_box_attack}, `Source' and `Target' represent the robust accuracy when adversarial samples generated by the gradient of the source method are transferred to the target method. Based on the results of Table \ref{tab:tras_white_box_attack}, we observe the following: \textbf{1)} LoRA exhibits the best robustness accuracy when facing adversarial samples from different sources; \textbf{2)} The transfer of adversarial samples between PET techniques of the same type significantly impacts robust accuracy, whereas the transfer between different types of PET techniques has a relatively smaller influence.

In general, LoRA is less affected by adversarial samples from other techniques. Under attacks from four different adversarial samples, LoRA maintains an average robustness accuracy of 50.75\%. In contrast, S-Ada. and P-Ada. perform slightly worse, with average robustness accuracies of 48.83\% and 47.45\%, respectively. This difference likely stems from the fact that both S-Ada. and P-Ada. belong to the same type of PET technique, where fine-tuning is achieved by inserting bottleneck modules into the pre-trained model. As a result, the adversarial samples generated by these two methods exhibit some similarity, leading to higher transferability between them. However, adversarial samples demonstrate lower transferability between different types of PET techniques. For instance, adversarial samples generated by S-Ada. achieve a robustness accuracy of 40.71\% when attacking P-Ada., but this increases to 49.14\% when attacking LoRA.

\subsection{Robustness and PET Size}
\label{sec:wba-ps}

To fully explore the impact of learnable parameters in PET on robustness accuracy, we conduct scaling experiments. As shown in Table \ref{tab:pet_scale}, we derive the following conclusions: \textbf{1)} PET techniques do not strictly follow the scaling law, meaning that increasing the number of learnable parameters does not necessarily lead to improved fine-tuning performance; \textbf{2)} Different PET methods exhibit varying sensitivity to changes in the number of learnable parameters. Specifically, VPT demonstrates the least sensitivity to parameter changes in terms of robustness, LoRA shows higher sensitivity, while S-Ada. and P-Ada. display moderate sensitivity.

As shown in Table \ref{tab:pet_scale}a, under FGSM and PGD attacks, the range of $\Delta$ for VPT is only 2.07\% and 1.52\%, respectively. In contrast, in Table \ref{tab:pet_scale}d, LoRA exhibits a range of $\Delta$ of 5.16\% and 5.34\% under FGSM and PGD attacks. This indicates that the robustness of LoRA is more susceptible to changes in the number of learnable parameters.

\begin{table*}[!t]
\caption{Scaling the number of learnable parameters introduced by PET and reporting the Mean Accuracy (\%) on VTAB-1K under white-box attacks. \textbf{Bold} indicates the maximum value and \underline{underline} represents the minimum value.}
\label{tab:pet_scale}
\centering
\subfloat[VPT. $p$ is the prompt length. \label{tab:vpt_scale}]{
\setlength{\tabcolsep}{1.0pt}
\scalebox{1}{
\begin{tabular}{l|ccccc}
\toprule
$p$ &Clean &FGSM &$\Delta$ & PGD&$\Delta$ \\
\midrule
10 &\underline{54.61}&\underline{23.27}&\underline{31.34}&\underline{20.69}&\underline{33.92}\\
 20 &59.52&26.68&32.84&25.31&34.21 \\
 50 &63.25&29.84&\textbf{33.41}&27.81&\textbf{35.44} \\
 100 &63.24&30.71&32.53&28.66&34.58 \\
 200 &\textbf{64.24}&\textbf{31.85}&32.39&\textbf{28.99}&35.25 \\
\bottomrule
\end{tabular}}
}
\subfloat[S-Ada. $d^{\prime}$ is the hidden size. \label{tab:sada_scale}]{
\setlength{\tabcolsep}{1.0pt}
\scalebox{1}{
\begin{tabular}{l|ccccc}
\toprule
$d^{\prime}$ &Clean &FGSM &$\Delta$ & PGD &$\Delta$ \\
\midrule
4 &\textbf{69.28}&\textbf{34.00}&\textbf{35.28}&\textbf{31.92}&\textbf{37.36} \\
8 &67.43&33.22&34.21&31.32&36.11 \\
16 &67.76&33.22&34.54&30.92&36.84 \\
32 &66.55&32.78&33.77&30.84&35.71 \\
64 &\underline{59.48}&\underline{28.41}&\underline{31.07}&\underline{25.82}&\underline{33.66} \\
\bottomrule
\end{tabular}}
}
\subfloat[P-Ada. $d^{\prime}$ is the hidden size. \label{tab:pada_scale}]{
\setlength{\tabcolsep}{1.0pt}
\scalebox{1}{
\begin{tabular}{l|ccccc}
\toprule
$d^{\prime}$ &Clean &FGSM &$\Delta$ & PGD &$\Delta$ \\
\midrule
4 &\underline{67.74}&\underline{32.88}&34.86& \underline{30.46}& 37.28\\
8 &68.72&33.52& \textbf{35.20} &31.00&\textbf{37.72} \\
16 &66.94&33.23&33.71&31.03&35.91 \\
32 &\textbf{69.34}&\textbf{35.47}&33.87&32.27&37.07 \\
64 &67.85&34.85&\underline{33.00} &\textbf{32.77}&\underline{35.08} \\
\bottomrule
\end{tabular}}
}
\subfloat[LoRA. $r$ is the hidden size. \label{tab:lora_scale}]{
\setlength{\tabcolsep}{1.0pt}
\scalebox{1}{
\begin{tabular}{l|ccccc}
\toprule
$r$ &Clean &FGSM &$\Delta$ & PGD &$\Delta$ \\
\midrule
4 &\textbf{68.13}&32.60&\textbf{35.53}&30.76&\textbf{37.37} \\
8 &67.72&33.09&34.63&32.03&35.69 \\
16 &\underline{63.40}&\underline{30.58}&32.82& \underline{29.31}&34.09\\
32 &67.26&34.49&32.77& 32.63&34.63\\
64 &65.58&\textbf{35.21}&\underline{30.37}& \textbf{33.55}&\underline{32.03}\\
\bottomrule
\end{tabular}}
}
\end{table*}

\section{Robustness to Information Perturbations}

\begin{figure*}[!t]
    \centering
\includegraphics[width=0.85\linewidth]{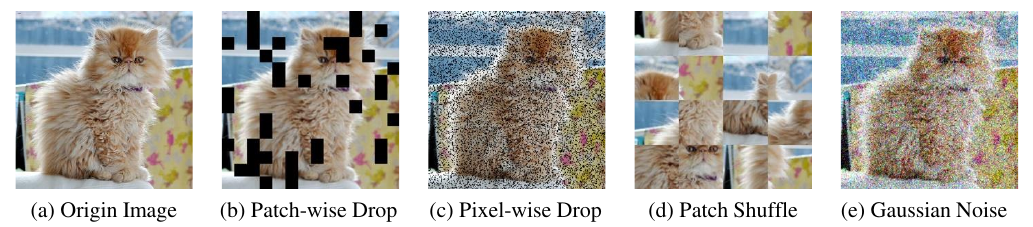}
    \caption{An example image with its various information perturbations.}
    \label{fig:examples}
\end{figure*}

\begin{figure*}[!t]
    \centering
\includegraphics[width=0.9\linewidth]{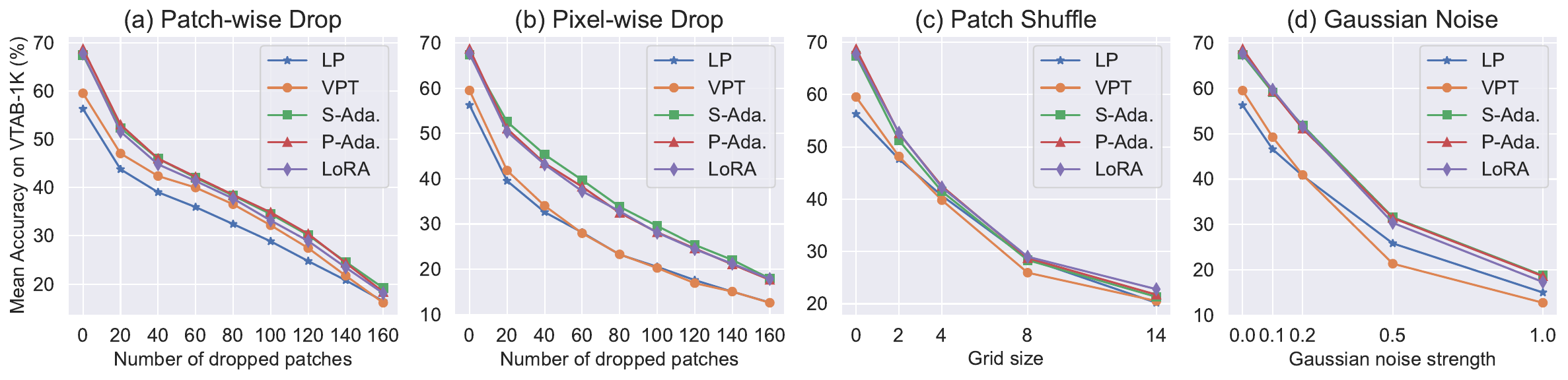}
    \caption{Mean Accuracy (\%) on VTAB-1K with various information perturbations.}
    \label{fig:info_per_res}
\end{figure*}

The ViT model \cite{vit,vitsurvey,vit22b} transforms images into sequences of patches as input. When parts of these sequences are missing, affected by noise, or their order is disrupted, the model's performance is influenced to some extent. To validate the changes in robustness after introducing PET techniques into ViT, we implement four types of information perturbation strategies. Specifically, as shown in Fig. \ref{fig:examples}, we apply Patch-wise Drop, Pixel-wise Drop, Patch Shuffle, and Gaussian Noise operations to the original images. As shown in Figure \ref{fig:info_per_res}, the following conclusions can be drawn: \textbf{1)} Patch-wise Drop and Patch Shuffle primarily interfere with the ViT architecture itself, with minimal relevance to specific fine-tuning methods; \textbf{2)} Compared to the dense information loss caused by Patch-wise Drop, the sparse information loss induced by Pixel-wise Drop has a more significant impact on VPT. This is because VPT fine-tunes through the introduction of learnable tokens, and sparse information loss disrupts each token, leading to a more pronounced effect on VPT; \textbf{3)} Similar to Pixel-wise Drop, Gaussian Noise essentially introduces additional noise to each token, thereby having a greater impact on VPT while affecting the other three PET methods to a lesser extent.

\subsection{Patch-wise Drop}

The Patch-wise Drop operation affects larger, contiguous areas of the image and is designed to assess the robustness of PET techniques when facing dense information loss. For instance, as shown in Figure \ref{fig:examples}(b), we discard 40 patches with a size of 16×16.

As shown in Figure\ref{fig:info_per_res}(a), it can be observed that when the number of discarded patches is 20, the robustness accuracy of the best-performing S-Ada. and the worst-performing VPT differs by 5.26\%. However, as the number of discarded patches increases, the performance of all PET methods declines. Specifically, when the number of discarded patches reaches 160, S-Ada. achieves a robustness accuracy of 19.19\%, while VPT achieves 16.13\%, narrowing the gap between them. Furthermore, overall, the performance of all fine-tuning methods is relatively close. This indicates that the perturbation caused by dense information dropout primarily targets the ViT structure itself, with minimal correlation to the PET methods.

\subsection{Pixel-wise Drop}

We introduce 8 different perturbation intensities in Pixel-wise Drop, ranging from 20 to 160. As shown in Figure \ref{fig:examples}(c), we randomly drop pixels sized 40×16×16. This operation distributes perturbations evenly across each patch, resulting in sparser perturbations and broader coverage.

As shown in Figure \ref{fig:info_per_res}(b), as the perturbation intensity increases, each patch is progressively affected, influencing every image token. This results in an increasingly noticeable performance gap between VPT and the other three PET methods. The reason lies in the fact that VPT introduces learnable tokens that interact with other image tokens to adjust the model's predictions for downstream tasks. Consequently, VPT is more affected by disturbed image tokens. In contrast, the other three PET methods, which introduce linear layers into the model, are less influenced by damaged image tokens.

\subsection{Patch Shuffle}

In the Patch Shuffle operation, we rearrange the order of patches within an image to evaluate the robustness after the reorganization of information. As shown in Figure \ref{fig:examples}(d), the image is shuffled based on a 4×4 grid. The results are presented in Figure \ref{fig:info_per_res}(c). Similar to Patch-wise Drop, Patch Shuffle introduces minimal differences across various PET techniques. In essence, this perturbation method primarily targets the ViT architecture itself, with little relation to the fine-tuning method used.

\subsection{Gaussian Noise}

We simulate the effect of environmental noise by applying Gaussian noise of varying intensities to the images. Figure \ref{fig:examples}(e) shows the image after adding Gaussian noise with an intensity of 0.2. As illustrated in Figure \ref{fig:info_per_res}(d), the performance gap between VPT and the other three PET methods widens as the noise intensity increases. Similar to the Pixel-wise Drop operation, Gaussian noise perturbation introduces additional noise to each pixel, affecting every image token, which leads to a more pronounced impact on VPT.

\section{Conclusions}
In this paper, we systematically investigate the robustness of four representative PET techniques: VPT, S-Ada., P-Ada., and LoRA. Through experiments on white-box attacks and information perturbations, we observe that while VPT demonstrates strong robustness under white-box attacks, its performance significantly lags behind the other three PET methods when facing information perturbation. Furthermore, we thoroughly explore the robustness variations of these PET methods with different learnable parameters and their performance under adversarial samples transfer. These comprehensive analyses enhance our understanding of the robustness of PET techniques and provide critical insights for their future improvement.

\bibliographystyle{IEEEtrans}
\bibliography{refs}

\end{document}